\documentclass{article}

\PassOptionsToPackage{numbers,sort&compress}{natbib}
\usepackage[sglblindworkshop,final]{neurips_2026}
\workshoptitle{AIMO Interpretability Challenge}

\usepackage[utf8]{inputenc}
\usepackage[T1]{fontenc}
\usepackage{hyperref}
\usepackage{url}
\usepackage{booktabs}
\usepackage{array}
\usepackage{amsfonts}
\usepackage{amsmath}
\usepackage{nicefrac}
\usepackage{microtype}
\usepackage{xcolor}
\usepackage{graphicx}
\usepackage{enumitem}
\usepackage{seqsplit}
\usepackage{float}
\usepackage{pifont}
\usepackage[font=small,labelfont=bf]{caption}
\usepackage{tikz}
\usetikzlibrary{arrows.meta}
\usepackage{algorithm}
\usepackage{algpseudocode}
\setcitestyle{square,comma}
\hypersetup{colorlinks=true,linkcolor=black,citecolor=black,urlcolor=black}
\setlist{nosep,leftmargin=1.2em}

\definecolor{ishraqcolor}{HTML}{228B22}
\definecolor{headerlink}{HTML}{367DBD}

\newcommand{\cmark}{\ding{51}}
\newcommand{\xmark}{\ding{55}}
\newcommand{\pfam}[1]{\textit{#1}}  

\title{Easy to Solve, Hard to Break?\\ Label-Free Prediction of Answer Robustness\\ in Reasoning Models}

\author{%
  \mdseries Md Farhan Ishraq$^{1}$, Didhiti Nahid$^{1}$, Nakib Saleh$^{1}$, Md Farhan Ishmam$^{2}$ \\
  $^{1}$Islamic University of Technology \quad $^{2}$University of Utah \\
  \texttt{farhanishraq17@iut-dhaka.edu}, \texttt{farhan.ishmam@utah.edu} \\
  \hypersetup{urlcolor=headerlink}\url{https://github.com/farhanishraq17/BakeNekos-AIMO-Smol-Interpretability} \\
  \hypersetup{urlcolor=headerlink}\url{https://github.com/farhanishraq17/BakeNekos-AIMO-Large-Interpretability}
}

\begin{document}

\maketitle

\begin{abstract}
A reasoning model can answer a competition problem correctly and still fail once the problem is reworded or given a
contradictory premise. The AIMO Interpretability Challenge asks if such fragility can be predicted from the model itself, with almost no labeled data. In the organizers' labeled split, models labeled on the same problem receive the same label, so robustness behaves as a property of the problem. We therefore propose the
\emph{difficulty probe}. Linear probes trained on 497 unlabeled olympiad problems read how long a reference model would reason about a problem, and each model's threshold follows in closed form from class balances. Trained without any robustness label, the probe improves over per-model
constants on both tracks, and an ablation credits the gain to the hidden states. On the 34 labeled
small-model rows, it is correct on 28, against 18 for constants and 24 for a length-only baseline. We also show why
richer approaches failed: labels are decided mostly by an added contradictory premise that models often fail to
notice, offline validation ranked candidate methods in the opposite order to the hidden set, and the
difficulty--robustness relation reverses on a second population.
\end{abstract}

\section{Introduction}

Reasoning models now solve much of competition mathematics, yet a correct answer is weak evidence of a correct
process. Accuracy falls when a problem's numbers change or an irrelevant clause is added~\citep{mirzadeh2024gsm}, on
perturbed competition benchmarks~\citep{huang2025mathperturb,gulati2025putnam}, and when a problem is made
unsolvable~\citep{xue2025reliablemath,fan2025missing}. The AIMO Interpretability
Challenge~\citep{aimo2026site,stefanik2026aimo} asks, for each model and problem, whether the answer is \emph{robust},
meaning that it survives answer-preserving rewrites, or \emph{spurious}.

A few properties make this task challenging. First, labeled data are scarce. The organizers' labeled split contains only 71 labeled rows across 36 problems~\citep{trainmainv2}, while the larger labeled source available to us, Hendrycks MATH~\citep{hendrycks2021math}, comes from a different distribution. Second, evaluation is constrained. The predictor must run offline on a single graphics processing unit (GPU), receiving one problem per call, with a limit of 160\,s per call in the final test~\citep{gettingstarted}. Third, as we show, the competition labels are determined primarily by a perturbation that rarely affects MATH labels.

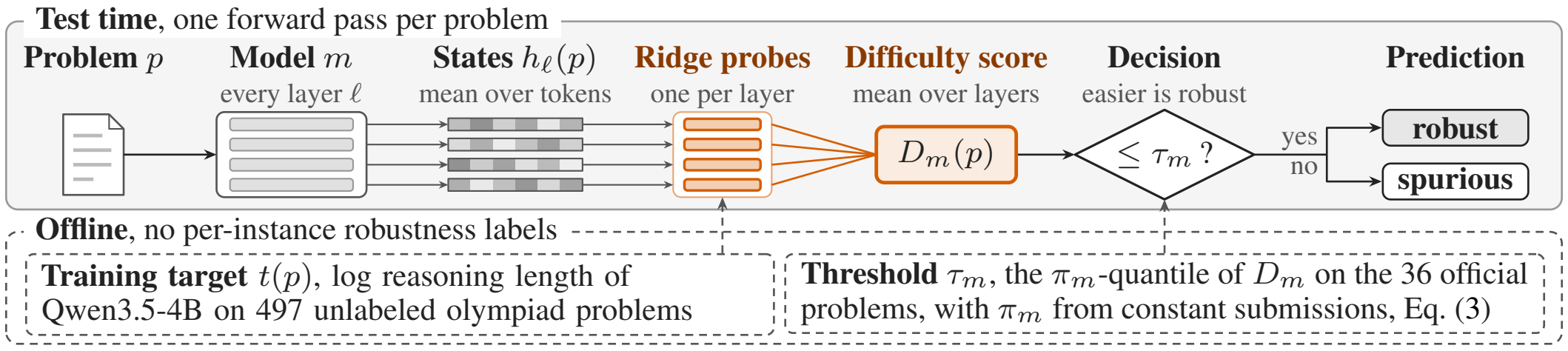
\begin{figure}[t]
\centering
\definecolor{fmProbe}{HTML}{D55E00}%
\definecolor{fmProbeDark}{HTML}{8A3A00}%
\definecolor{fmInk}{HTML}{222222}%
\definecolor{fmDark}{HTML}{555555}%
\definecolor{fmMid}{HTML}{999999}%
\definecolor{fmLight}{HTML}{BBBBBB}%
\begin{tikzpicture}[x=1cm, y=1cm, font=\fontsize{8}{9}\selectfont, line join=round,
  >={Stealth[length=3.2pt, width=2.6pt]},
  flow/.style={->, draw=fmInk, line width=0.5pt},
  thin flow/.style={->, draw=fmDark, line width=0.4pt, >={Stealth[length=2.4pt, width=2pt]}},
  offl/.style={->, draw=fmDark, line width=0.5pt, dash pattern=on 2pt off 1.5pt},
  head/.style={align=center, text=fmInk, inner sep=0pt, anchor=north},
  lane/.style={fill=white, inner xsep=3pt, inner ysep=0.5pt, text=fmInk, anchor=west},
  offbox/.style={draw=fmDark, dash pattern=on 2pt off 1.5pt, line width=0.5pt, fill=white, rounded corners=2pt,
    align=left, inner xsep=4pt, inner ysep=2pt, text=fmInk, font=\fontsize{8}{8.8}\selectfont},
  slab/.style={draw=fmMid, fill=fmLight!45, line width=0.4pt, rounded corners=0.8pt},
  pchip/.style={draw=fmProbe, fill=fmProbe!30, line width=0.65pt, rounded corners=0.8pt},
  tag/.style={draw=fmInk, line width=0.5pt, rounded corners=2pt, minimum width=1.30cm, minimum height=0.32cm,
    inner sep=1pt, text=fmInk}]

\def\rows{0.22,0.40,0.58,0.76}
\def\yc{0.49}
\def\hs{\vphantom{$h_\ell(p)$}}
\def\hy{1.49}
\def\ytop{1.68}
\def\yoff{-0.20}
\def\ybot{-1.20}

\draw[fmMid, fill=fmLight!14, line width=0.5pt, rounded corners=3pt] (0,0) rectangle (13.9,\ytop);
\draw[fmDark, line width=0.5pt, dash pattern=on 2pt off 1.5pt, rounded corners=3pt] (0,\ybot) rectangle (13.9,\yoff);
\node[lane] at (0.15,\ytop) {\textbf{Test time}, one forward pass per problem};
\node[lane] at (0.15,\yoff) {\textbf{Offline}, no per-instance robustness labels};

\node[head] at (0.78,\hy) {\hs\textbf{Problem} $p$};
\node[head] at (2.55,\hy) {\hs\textbf{Model} $m$\\ {\scriptsize\color{fmDark} every layer $\ell$}};
\node[head] at (4.55,\hy) {\hs\textbf{States} $h_\ell(p)$\\ {\scriptsize\color{fmDark} mean over tokens}};
\node[head] at (6.40,\hy) {\hs\textcolor{fmProbeDark}{\textbf{Ridge probes}}\\ {\scriptsize\color{fmDark} one per layer}};
\node[head] at (8.40,\hy) {\hs\textcolor{fmProbeDark}{\textbf{Difficulty score}}\\ {\scriptsize\color{fmDark} mean over layers}};
\node[head] at (10.35,\hy) {\hs\textbf{Decision}\\ {\scriptsize\color{fmDark} easier is robust}};
\node[head] at (12.95,\hy) {\hs\textbf{Prediction}};

\begin{scope}[shift={(0.78,\yc)}]
  \draw[fmDark, fill=white, line width=0.5pt] (-0.27,-0.36) -- (-0.27,0.36) -- (0.14,0.36) -- (0.27,0.23)
    -- (0.27,-0.36) -- cycle;
  \draw[fmDark, line width=0.4pt] (0.14,0.36) -- (0.14,0.23) -- (0.27,0.23);
  \foreach \yy/\xx in {0.17/0.08, 0.04/0.16, -0.09/0.16, -0.22/0.00}
    \draw[fmLight, line width=0.9pt] (-0.18,\yy) -- (\xx,\yy);
\end{scope}

\draw[fmDark, line width=0.5pt, rounded corners=2pt, fill=white] (1.88,0.11) rectangle (3.22,0.87);
\foreach \y in \rows
  \draw[slab] (2.00,\y-0.055) rectangle (3.10,\y+0.055);
\draw[flow] (1.05,\yc) -- (1.88,\yc);

\draw[fmProbe!55, line width=0.4pt, rounded corners=2pt, fill=white] (5.96,0.11) rectangle (6.84,0.87);
\coordinate (pgroup) at (6.40,0.11);

\foreach \y [count=\i] in \rows {
  \draw[thin flow] (3.22,\y) -- (3.95,\y);
  \foreach \c/\s in {0/22, 1/48, 2/12, 3/36, 4/58, 5/28} {
    \pgfmathsetmacro{\shade}{mod(\s + 17*\i, 60) + 8}
    \fill[fmDark!\shade] (3.95+0.2*\c,\y-0.055) rectangle (4.15+0.2*\c,\y+0.055);
  }
  \draw[fmDark, line width=0.4pt] (3.95,\y-0.055) rectangle (5.15,\y+0.055);
  \draw[thin flow] (5.15,\y) -- (5.96,\y);
  \draw[pchip] (6.06,\y-0.05) rectangle (6.74,\y+0.05);
  \draw[fmProbe, line width=0.5pt] (6.84,\y) -- (7.775,\yc);
}

\node[draw=fmProbe, fill=fmProbe!12, line width=0.9pt, rounded corners=2pt, minimum width=1.25cm,
  minimum height=0.50cm, inner sep=1pt, text=fmInk, font=\footnotesize] (D) at (8.40,\yc) {$D_m(p)$};

\coordinate (dc) at (10.35,\yc);
\draw[fmInk, fill=white, line width=0.6pt]
  (10.35-0.80,\yc) -- (10.35,\yc+0.40) -- (10.35+0.80,\yc) -- (10.35,\yc-0.40) -- cycle;
\node[text=fmInk, font=\footnotesize, inner sep=0pt] at (dc) {$\le\tau_m$\,?};
\draw[flow] (D.east) -- (10.35-0.80,\yc);

\node[tag, fill=fmLight!35] (rob) at (12.95,0.73) {\textbf{robust}};
\node[tag, fill=white] (spu) at (12.95,0.25) {\textbf{spurious}};
\draw[fmInk, line width=0.5pt] (10.35+0.80,\yc) -- (11.80,\yc);
\draw[flow] (11.80,\yc) |- node[pos=0.25, left, inner sep=2pt, text=fmDark, font=\scriptsize] {yes} (rob.west);
\draw[flow] (11.80,\yc) |- node[pos=0.25, left, inner sep=2pt, text=fmDark, font=\scriptsize] {no} (spu.west);

\node[offbox, anchor=north west, text width=6.40cm] (tgt) at (0.17,-0.39)
  {\textbf{Training target} $t(p)$, log reasoning length of\\
   Qwen3.5-4B on 497 unlabeled olympiad problems};
\node[offbox, anchor=north east] (thr) at (13.75,-0.39)
  {\textbf{Threshold} $\tau_m$, the $\pi_m$-quantile of $D_m$ on the 36 official\\
   problems, with $\pi_m$ from constant submissions, Eq.~\eqref{eq:share}};
\draw[offl] (tgt.north -| pgroup) -- (pgroup);
\draw[offl] (thr.north -| dc) -- (10.35,\yc-0.40);
\end{tikzpicture}%
\caption{\textbf{The difficulty probe.} At test time (solid), ridge probes read model $m$'s prompt states in one forward pass,
and their mean $D_m(p)$ is compared with a threshold $\tau_m$. Offline (dashed), they fit a reference model's log
reasoning length without any robustness label, and class balances set $\tau_m$.}
\label{fig:method}
\end{figure}

Whenever two evaluated models are labeled on the same official problem, the labels agree, in 20 of 20 pairs among the
four open models and on 11 of 11 problems once gpt-oss-120b is included. Robustness therefore behaves as a property of
the problem, so a problem-level quantity that needs no robustness labels can predict it. We use difficulty, a
reference model's reasoning length, and read it from the scored model's prompt representations with linear probes in
one forward pass (Figure~\ref{fig:method}). This \emph{difficulty probe} is our submission to both tracks. Our contributions are as follows:
\begin{itemize}
\item A robustness predictor, trained without robustness labels, that beats constant and text-only baselines on
  both tracks, and a pre-registered ablation crediting the gain to the hidden states (Table~\ref{tab:results}).
\item Exploratory evidence that the probe reads difficulty spread over many directions, and a pre-registered test in which, at the tested dose, steering shifts the probe's reading but not the model's reasoning length (Figure~\ref{fig:reads}).
\item An added contradictory premise determines most competition labels, while a separate perturbation family, \emph{domain}, determines most MATH labels, which explains why predictors trained on MATH labels do not transfer (Section~\ref{sec:deciding}).
\item Twenty negative results, accompanied by an assessment of their statistical power, and three validation failures that may recur in similar settings (Section~\ref{sec:negatives}). 

\end{itemize}

\section{Methodology}

\subsection{Problem setting}
\label{sec:setting}

For a model $m$ and problem $p$, let $a_0 \in [0,1]$ be the accuracy of $m$ on $p$ and $a_\phi \in [0,1]$ its accuracy
on rewrites from perturbation family $\phi \in \{$\pfam{rephrase}, \pfam{rename}, \pfam{domain}, \pfam{distract},
\pfam{typos}, \pfam{expert perturbations}, \pfam{expert no-solution}$\}$. Every family preserves the answer except
\pfam{expert no-solution}, which adds a premise contradicting the rest of the problem, so the correct response is to
report that no solution exists. Pairs with $a_0 = 1$ are labeled $y(m,p) \in \{\text{robust}, \text{spurious}\}$ by
their largest accuracy drop $\Delta(m,p)$,
\begin{equation}
\Delta(m,p) = \max_{\phi}\,\bigl(a_0 - a_\phi\bigr), \qquad
y(m,p) = \begin{cases} \text{robust} & \text{if } \Delta(m,p) \le 0.10,\\
\text{spurious} & \text{if } \Delta(m,p) \ge 0.25,\end{cases}
\label{eq:label}
\end{equation}
and pairs in between are left unlabeled, a rule we verified on all 82 published rows. The Small Models Track (Small
Track) evaluates Qwen3.5-4B, Skywork-OR1-Math-7B, Olmo-3-7B-Think and DeepSeek-R1-0528-Qwen3-8B (Qwen, Skywork, Olmo
and DeepSeek below), and the Main Track adds gpt-oss-120b at three reasoning efforts. The organizers' labeled split
(the \emph{official split}) has 34 labeled small-model rows and 37 gpt-oss-120b rows, and we also use DeepSeek's 137
labeled Hendrycks MATH rows (the \emph{MATH split}), labeled under an older rule. Hidden validation sets of 14
(Small) and 21 (Main) cases form the leaderboards, and our nominated submissions are Codabench 1010961 (Small) and
1010444 (Main).

\subsection{The difficulty probe}
\label{sec:method}

\paragraph{Reading difficulty.} The difficulty $t(p) \in \mathbb{R}$ of a problem $p$ is the logarithm of the number of tokens Qwen generates for it,
capped at 32{,}768 and averaged over two samples drawn with the organizers' decoding settings. We compute it, without
using correctness, on 497 olympiad problems from OlympiadBench~\citep{he2024olympiadbench}, AMC and AIME that are
disjoint from the official split. For each scored model $m$, one forward pass over $p$, wrapped in the challenge's system
prompt and $m$'s chat template, gives a mean-pooled prompt state $h_\ell(p)$, a real vector, at each layer $\ell \in
\{0, \dots, L_m\}$. Per-layer ridge regressions with weights $w_{m\ell}$ and bias $b_{m\ell} \in \mathbb{R}$ predict
$t$ from states standardized by their mean $\mu_{m\ell}$ and standard deviation $\sigma_{m\ell}$. The difficulty score
$D_m(p) \in \mathbb{R}$ averages the layers:
\begin{equation}
D_m(p) = \frac{1}{L_m+1}\sum_{\ell=0}^{L_m}\Bigl(w_{m\ell}^{\top}\,\frac{h_\ell(p)-\mu_{m\ell}}{\sigma_{m\ell}}
+ b_{m\ell}\Bigr).
\label{eq:probe}
\end{equation}
All four models share the target, so each reads the same notion of difficulty from its own representations
(cross-validated Spearman correlation of 0.77--0.79 between predicted and observed difficulty).

\paragraph{Thresholds from class balances.} We call easier problems robust, predicting robust when $D_m(p) \le \tau_m$ for a threshold $\tau_m \in \mathbb{R}$.
This direction came from an exploratory analysis of the official split, where shorter reasoning traces accompanied
robust labels, with an area under the receiver operating characteristic curve (AUC) of 0.71 [0.45, 0.94].
Section~\ref{sec:transfer} therefore tests it on a second population. Let $r_m$ and $q_m$ count model $m$'s robust and
spurious hidden validation cases, with $n_m = r_m + q_m$ and $d_m = r_m - q_m$. Predicting robust for $m$'s cases and spurious for all others scores
$d_m$ cases more than predicting spurious everywhere, so constant submissions measure $d_m$ and give $m$'s robust
share $\pi_m \in [0,1]$,
\begin{equation}
\pi_m = \frac{r_m}{n_m} = \frac{n_m + d_m}{2\,n_m}.
\label{eq:share}
\end{equation}
We set $\tau_m$ to the $\pi_m$-quantile of $D_m$ over the 36 official problems, which share the evaluation sets'
pipeline. Constant submissions gave $d = (+1, +1, -1, -1)$ for Qwen, DeepSeek, Skywork and Olmo, and scored
submissions imply $n = (5, 3, 3, 3)$, hence $\pi = (0.6, \nicefrac{2}{3}, \nicefrac{1}{3}, \nicefrac{1}{3})$.

\paragraph{The Main Track.} gpt-oss-120b cannot be loaded in the published evaluation image, where its MXFP4 checkpoint dequantizes to 217.6\,GiB
against 95\,GiB of GPU memory. Its 11 of the 21 Main Track cases therefore receive one constant per reasoning effort,
chosen with constant submissions, and no white-box signal.

\section{Results}
\label{sec:experiments}

We ask five research questions (RQs). Does the probe beat label-free baselines because of its hidden states
(\textbf{RQ1})? Does the relation hold on another population (\textbf{RQ2})? What does the probe read, and does the
model use it (\textbf{RQ3})? What decides the label (\textbf{RQ4})? What do our failures teach (\textbf{RQ5})? As the
hidden validation sets are small, we also report accuracy and AUC on the official split's 34 small-model rows, with
95\% intervals from a bootstrap over problem templates, which several problems share. The baselines are per-model
constants (every method's fallback), problem length in characters, and familiarity, the mean token negative
log-likelihood (NLL) of the problem under the scored model, a membership-inference
statistic~\citep{shi2023detecting,carlini2022membership}. Length uses the probe's calibration, and familiarity was
tried with three calibrations, including the probe's. All gates and comparisons were pre-registered unless marked
exploratory.

\begin{table}[t]
\caption{\textbf{Per-instance results (RQ1).} Hidden validation: cases correct, one problem per call. Official split: 34
small-model rows, 95\% template-bootstrap intervals (familiarity's balanced accuracy is 0.33--0.46). Main: all
methods share gpt-oss-120b's constants and differ only in the small models' signal. WB: uses hidden states.}
\label{tab:results}
\centering\small
\setlength{\tabcolsep}{5pt}
\begin{tabular}{@{}lccccc@{}}
\toprule
& & \multicolumn{2}{c}{Hidden validation} & \multicolumn{2}{c}{Official split}\\
\cmidrule(lr){3-4}\cmidrule(l){5-6}
Method & WB & Small (14) & Main (21) & Accuracy & AUC\\
\midrule
Constant per model & \xmark & 9 & 12 & 0.53 & --\\
Problem length & \xmark & -- & 12 & 0.71 & 0.53 {\scriptsize [0.22, 0.94]}\\
Familiarity (token NLL) & \xmark & 8--9 & 13 & -- & --\\
\textbf{Difficulty probe (ours)} & \cmark & \textbf{10} & \textbf{14} & \textbf{0.82} & \textbf{0.88} {\scriptsize [0.71, 1.00]}\\
\bottomrule
\end{tabular}
\end{table}

\subsection{Main results and white-box ablation (RQ1)}
\label{sec:main}

The difficulty probe is the best per-instance method on every measure in Table~\ref{tab:results}. It beats the
constants by one Small and two Main Track cases, whereas familiarity does not on the Small Track. As one or two cases
are weak evidence, the official split gives a second test, where the probe's accuracy exceeds the constants' by 0.29
[0.06, 0.52] and problem length's by 0.12 [0.02, 0.26]. Only the input signal differs between these methods, so the comparisons form a white-box ablation. On the Main Track hidden set, both predictions that the probe changes are
correct, and the run took 59.9\,s including model loading. The ablation thus credits the gain to the hidden states,
although the official split, which informed the probe's direction, is not a clean held-out test.

\subsection{The relation depends on the population (RQ2)}
\label{sec:transfer}

\begin{figure}[t]
\centering
\includegraphics[width=\linewidth]{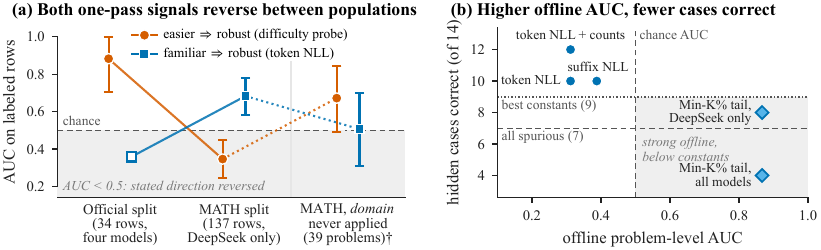}
\caption{\textbf{(a) Both one-pass signals reverse between populations.} AUC (95\% intervals) of ``easier $\Rightarrow$
robust'' and ``familiar $\Rightarrow$ robust'' ($\dagger$: selected post hoc, hollow: no recorded interval).
\textbf{(b) Offline validation ranked methods in the opposite order to the hidden set.} Offline AUC against hidden cases correct for familiarity
statistics (Small Track, batch calls).}
\label{fig:direction}
\end{figure}

On the official split, easier problems are more robust and familiar ones less so, and on the MATH split both reverse
(Figure~\ref{fig:direction}a). Measured by AUC, easier $\Rightarrow$ robust (probe) falls from 0.883 [0.706, 1.000] on the official split to 0.346 [0.246, 0.447] on the MATH split, and familiar $\Rightarrow$ robust (token NLL) rises from 0.358 to 0.683 [0.582, 0.780] (Table~\ref{tab:fig2a}). The signals are linked, as harder problems are more familiar to every model (Spearman
$-0.35$ to $-0.50$), so neither direction holds on both populations. An exploratory analysis attributes most of the
reversal to one family. On MATH, \pfam{domain} decides most labels and was applied more often to easy problems (88\% of
level-1 but 45\% of level-5 problems), and on the 39 problems it never touched, the official pattern returns for
difficulty (Table~\ref{tab:fig2a}). Because this subset was chosen after the reversal was observed, it offers a
plausible explanation rather than evidence that the probe transfers.

\subsection{What the probe reads (RQ3)}
\label{sec:reads}

\begin{figure}[t]
\centering
\includegraphics[width=\linewidth]{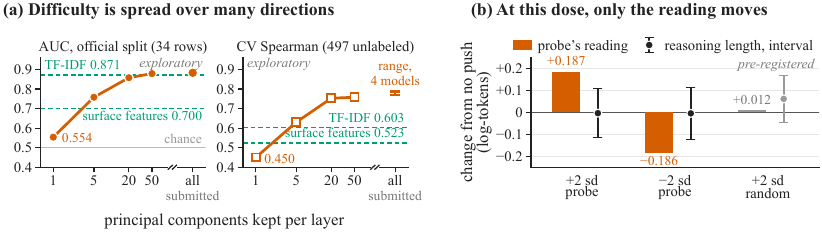}
\caption{\textbf{(a) Difficulty is spread over many directions (exploratory).} Official-split AUC and cross-validated Spearman
on the 497 unlabeled problems, for probes on the top 1, 5, 20 or 50 principal components per layer or on all
dimensions (submitted). TF-IDF: term frequency--inverse document frequency. \textbf{(b) At this dose, steering moves
the probe's reading but not reasoning length (pre-registered).} Qwen, layer 15, 60 unlabeled problems.}
\label{fig:reads}
\end{figure}

Exploratory analyses suggest that the probe reads a distributed property of the problem. Difficulty is decodable at
every layer but needs many directions (Figure~\ref{fig:reads}a). The four models read it alike (pairwise Spearman
0.926 or higher), and surface features explain about half of the variance in the probe's reading (Appendix~\ref{app:results}). Word TF-IDF features
rank the official split almost as well but predict held-out difficulty much worse, so only the 497 unlabeled problems
separate the two, in favor of the hidden states. Refitting the probe on random subsets of its training problems changes
its official-split AUC by about 0.08 (0.815 $\pm$ 0.082 with 125 problems), so the official split cannot distinguish
refits of one method. In a pre-registered test of whether the model uses this direction,
pushing Qwen's prompt states two standard deviations (sd) along it at layer 15 moved the probe's reading of later
layers by about $\pm0.19$ log-tokens, but left the model's reasoning length unchanged (Figure~\ref{fig:reads}b).


\subsection{What decides the label (RQ4)}
\label{sec:failures}
\label{sec:deciding}

A pair is spurious when any one family lowers accuracy by at least 0.25 (Equation~\ref{eq:label}), so the decisive family
determines what a predictor must learn. On solved MATH rows, \pfam{domain} rewrites lower DeepSeek's accuracy by that
much on 86\% of rows. In the organizers' olympiad sample, the decisive family is instead \pfam{expert no-solution}
(43\% across eight systems), and surface families rarely matter (10\% or less), although these estimates rest on seven
eligible olympiad problems. A predictor trained on MATH labels therefore learns a sensitivity that seldom changes
olympiad labels, consistent with all such predictors we built scoring near chance there (AUC 0.43--0.62). On 164
human-verified contradiction rewrites of AMC problems~\citep{xue2025reliablemath} with a 16k-token budget, Skywork
flagged the contradiction in none of 274 samples and returned the original answer in 44\% of them. Qwen exhausted its
budget on 79\% of samples instead of declining, as reported for missing and flawed
premises~\citep{fan2025missing,li2025premisecritique}. In a second pre-registered study, Qwen flagged inserted
contradictions less often on problems it found familiar (Spearman 0.256 [0.112, 0.386] between token NLL and
detection rate), although problem length explains part of this association (0.142 [$-0.015$, 0.282] within length
terciles, Appendix~\ref{app:contradiction}). Labels on competition problems thus largely record whether a model checks its
premises, which none of our representation-based methods was designed to capture.

\subsection{Why richer approaches failed (RQ5)}
\label{sec:negatives}

Appendix~\ref{app:negatives} lists twenty rejected hypotheses, several of which reached an AUC of 0.73--0.87 before
the test that rejected them. One concerns the challenge's probe baseline, which reports 58.4\% accuracy although its
fits average 0.501 per layer and nested layer selection gives 0.552, a selection effect of the kind that control tasks
expose~\citep{hewitt2019control}. Another concerns counterfactual prompt-state sensitivity, which reached an AUC of 0.735 but only 0.545 [0.428, 0.658]
on an independent population (Table~\ref{tab:negatives}, row 9).
Labels we generated with the organizers' own procedure failed our pre-registered
quality gates (Appendix~\ref{app:labels}). Most of our tests were underpowered, because detecting the effects we
observed at 80\% power required 63--320 labeled rows but only 33--61 were available (power 14--53\%), so these null
results bound effects only loosely.

\label{sec:validation}
Under batch evaluation (before 15 October), the strongest offline candidate, a Min-K\% tail statistic~\citep{shi2023detecting,zhang2024minkpp}, achieved an offline AUC of 0.867 (maxT-adjusted $p = 0.001$~\citep{westfall1993resampling}) and predicted 4 of the 14 hidden cases correctly. Mean token NLL, despite failing its offline gate, predicted 10 correctly, or 12 when each model's number of robust predictions was set from its class balance (Figure~\ref{fig:direction}b). However, this approach ranked problems within each model's batch. Under per-instance evaluation, it therefore reduced to constant predictions, which score 9 of 14 Small cases and 12 of 21 Main Track cases (down from 15 of 21 under batch evaluation).

\section{Related work}
\label{sec:related}

\textbf{Perturbation studies} measure the fragility of mathematical
reasoning~\citep{mirzadeh2024gsm,huang2025mathperturb,gulati2025putnam,xue2025reliablemath,fan2025missing}, whereas we
predict it per problem from the model. \textbf{Linear probes} on a question's activations predict a reasoning model's
token count~\citep{sheng2025reasoningstrength}, and a language model's problem
difficulty~\citep{lee2025difficultyperception} and answer correctness~\citep{cencerrado2025noanswer}. Probing
results depend on capacity and selection~\citep{alain2016probes,hewitt2019control,belinkov2022probing}. We predict
robustness from a label-free difficulty read-out, and unlike the length direction of
\citet{sheng2025reasoningstrength}, ours did not change reasoning length at the dose we tested. Our familiarity
baseline, mean token NLL, is a \textbf{membership-inference
statistic}~\citep{shi2023detecting,zhang2024minkpp,carlini2021extracting,carlini2022membership}. \textbf{Uncertainty and self-consistency
methods}~\citep{kadavath2022know,kuhn2023semantic,farquhar2024semantic,wang2023selfconsistency} require sampling,
which a per-call budget makes costly, whereas our probe needs one forward pass.

\section{Conclusion}

A difficulty probe trained without robustness labels beat constant and text-only baselines on both tracks, and its
relation holds on competition problems but reverses on MATH. Our analysis suggests that competition labels mostly
record whether a model checks an added premise. Three lessons from our failures apply more broadly: decision rules
should not depend on how calls are grouped, methods should be validated on a population other than the one that chose
them, and compute cost should be checked against the budget before testing any signal (Table~\ref{tab:negatives}, rows 11 and 17).


\label{body-end}

\section*{Limitations}

\textbf{Sample size.} Our evidence comes from small samples. The hidden validation sets contain 14 and 21
cases, so a single case changes accuracy by 7.1 and 4.8 points. The official split contains 34 labeled
small-model rows and 37 gpt-oss-120b rows. Most of our negative results were obtained at 14--53\% power. The
private test set consists of 30--50 new problems, and its class balance may differ from that of the
validation sets.

\textbf{Selection.} The direction of the probe was chosen from an exploratory analysis of the official split,
which was later used to evaluate it. The MATH subset on which that direction is restored was selected after
the reversal was observed. The per-model shares come from the hidden validation set. The official split makes
Skywork-OR1-Math-7B and Olmo-3-7B-Think majority-robust (0.67 and 0.78), whereas the validation balances make
both spurious-leaning. If the test set resembles the official split, their thresholds are too strict. On the
Main Track, the per-effort constants for gpt-oss-120b were chosen with constant submissions on the same 21
cases that score them. The low-effort constant disagrees with the official split, where it is correct on 2
of 7 rows and always predicting robust is correct on 5.

\textbf{Scope of the white-box claim.} The claim covers the four small models only. The predictions for
gpt-oss-120b depend on the reasoning effort alone. The ablation compares a single black-box surrogate, problem
length, and has no shuffled-probe control. The causal test used one layer, one direction, one dose and one
model. The probe, however, reads a 20--50-dimensional subspace, so larger or subspace-wide interventions remain
untested.

\textbf{Decisions and environment.} For the Main Track, a pre-registered rule selected the variant that applies
Qwen3.5-4B's probe to gpt-oss-120b's problems. That variant also scored 14 of 21. We nominated the simpler variant
instead, after both results were known, because it ran faster (59.9\,s against 122.2\,s), the pre-registered variant showed
no measured gain, and the nominated run loads only the model being scored. Neither submission was tested in the
competition's Docker image. Their Codabench runs, with full coverage and no invalid predictions, are the
evidence that they run in the evaluation environment. Our finding that gpt-oss-120b cannot be loaded comes from
the published image, while the challenge's starter kit states that the model fits.

\textbf{Measurement of the label mechanism.} The per-family analysis rests on seven eligible olympiad problems.
The contradiction study used public American Mathematics Competitions (AMC) rewrites rather than the
organizers' expert rewrites, two of the four
models, and a 16k-token budget, whereas the organizers sample up to 100k tokens. Our own Main Track labels failed
both pre-registered quality gates and are used by no nominated submission. Their human verification (50
variants) was not completed. Finally, we did not pursue sparse
autoencoders~\citep{cunningham2023sae,bricken2023monosemanticity,templeton2024scaling}, because with 34 labeled rows,
a search over thousands of features has little statistical power, and our template-level bootstrap would reject
most of what it found.

\section*{AI Use Statement}

\textbf{Language models as experimental components.} The five evaluated models produced every hidden state,
likelihood and reasoning trace analyzed in this report. In the Main Track labeling program,
Kimi-K2-Instruct-0905 generated candidate rewrites, and Qwen3.5-397B-A17B-FP8, DeepSeek-R1-0528 and
GLM-5.1-FP8 verified them by re-solving each one. Contradiction variants were generated and checked by other
models: \texttt{nemotron-3-ultra-550b-a55b} and \texttt{minimax-m3.1-flash}, accessed through OpenRouter, in
one study, and Qwen3-8B-AWQ and DeepSeek-R1-Distill-Llama-8B in another. Their outputs entered our results only
through the scripted grading and checks described with each experiment, and through hand audits where a
protocol required one (a grader audit that agreed on 30 of 30 samples and a variant audit that found 29 of 30
genuine).

\textbf{Language models as research and writing assistants.} We used Claude Code, Anthropic's agentic coding
tool, to write and debug analysis, pipeline and packaging code, to draft pre-registrations, experiment logs and
documentation, and to edit this report.


\bibliographystyle{plainnat}
{\small \bibliography{references}}

\clearpage
\appendix

\section{Implementation and reproducibility}
\label{app:method}

\subsection{The predictor}

Algorithm~\ref{alg:probe} gives the complete inference procedure of both submissions. The Small Models Track
submission is the same procedure without the gpt-oss-120b branch. The per-model constant $c_m$ is
\texttt{True} (robust) for Qwen3.5-4B and DeepSeek-R1-0528-Qwen3-8B, the models with a positive class
balance. It is \texttt{False} for Skywork-OR1-Math-7B, Olmo-3-7B-Think and any unknown model. Because any exception returns $c_m$ (Algorithm~\ref{alg:probe}), a failed call returns the constant baseline's prediction.

\begin{algorithm}[H]
\caption{Difficulty probe, as submitted (one problem per call).}
\label{alg:probe}
\begin{algorithmic}[1]
\Require model identifier $m$, problems $P$, reasoning effort $e$
\If{$m$ is gpt-oss-120b} \Comment{Main Track, the model is never loaded}
  \State \Return $[\,e \ne \text{low}\,]$ for each $p \in P$
\EndIf
\If{no probe arrays exist for $m$} \Return $[c_m]$ for each $p \in P$ \EndIf
\State load $\{\mu_{m\ell}, \sigma_{m\ell}, w_{m\ell}, b_{m\ell}\}_\ell$ and $\tau_m$ \Comment{plain NumPy arrays}
\State load $m$ and its tokenizer, or reuse them if the previous call loaded $m$
\ForAll{$p \in P$}
  \State $x \gets$ chat template of $m$ applied to (baseline system prompt, $p$), truncated to 2{,}048 tokens
  \State $h_\ell \gets$ mean of the hidden states of $x$ at layer $\ell$, for $\ell = 0, \dots, L_m$
  \State $D \gets \frac{1}{L_m+1}\sum_\ell \bigl(w_{m\ell}^\top (h_\ell - \mu_{m\ell})/\sigma_{m\ell} + b_{m\ell}\bigr)$
  \State $\hat y_p \gets [\,D \le \tau_m\,]$
\EndFor
\State \Return $(\hat y_p)_{p \in P}$ \Comment{any exception returns $[c_m]$ for each $p$}
\end{algorithmic}
\end{algorithm}


\subsection{Training details}

\begin{table}[H]
\centering\small
\caption{Per-model fit and calibration. CV: five-fold cross-validated Spearman correlation between predicted
and observed difficulty. NLL: Spearman correlation between predicted difficulty and mean token
negative log-likelihood (NLL) over 548 olympiad problems. Layers: the number $L_m+1$ of layers averaged in Equation~\ref{eq:probe}. $d_m$, $n_m$ and $\pi_m$: the class balance, number of hidden validation cases and robust share of model $m$ on the Small Models Track (Section~\ref{sec:method}). $\tau_m$: the $\pi_m$-quantile of $D_m$ over the 36 official problems.}
\label{tab:fit}
\begin{tabular}{@{}lccccccc@{}}
\toprule
Model & Layers & $d_m$ & $n_m$ & $\pi_m$ & $\tau_m$ & CV & NLL\\
\midrule
Qwen3.5-4B & 33 & $+1$ & 5 & 0.6 & 10.3858 & 0.786 & $-0.351$\\
Skywork-OR1-Math-7B & 29 & $-1$ & 3 & $\nicefrac13$ & 9.9695 & 0.771 & $-0.504$\\
Olmo-3-7B-Think & 33 & $-1$ & 3 & $\nicefrac13$ & 9.9438 & 0.788 & $-0.416$\\
DeepSeek-R1-0528-Qwen3-8B & 37 & $+1$ & 3 & $\nicefrac23$ & 10.3925 & 0.768 & $-0.404$\\
\bottomrule
\end{tabular}
\end{table}

The 497 training problems come from OlympiadBench~\citep{he2024olympiadbench}, the American Mathematics
Competitions (AMC) and the American Invitational Mathematics Examination (AIME). They have reference answers and
exclude every problem of the official split. Qwen3.5-4B sampled each problem twice. The settings were temperature
1.0, top-$k$ 40, top-$p$ 0.95, a prompt that permits answering that no solution exists, and a cap of 32{,}768
tokens. Sampling used vLLM on pairs of T4 graphics processing units (GPUs) in fp16. The target $t(p)$ is censored
at the cap. Of the 497 problems, 194 have at least one capped sample and 141 have two.
The states of each layer are standardized, and a ridge regression is fitted to them. The penalty is chosen among nine values from $10^1$ to
$10^5$ by efficient leave-one-out cross-validation, and each regression is then refitted on all 497 problems. No
robustness label enters the fit. Labels influence only the direction of the decision (Section~\ref{sec:method})
and, through the class balances, the shares $\pi_m$.

\subsection{Submitted artifacts}

Both submissions ship the per-layer arrays as compressed NumPy files, the system prompt, the decision code and a
tokenizer guard that checks an exact round trip of the input. They contain no training data, labels or model weights.
\begin{itemize}
\item \textbf{Small Models Track} (Codabench 1010961): the submission archive (5.2\,MB). Before the archive is written, the shipped arithmetic is recomputed on the stored states of all 548 problems for all four models, and any deviation of $10^{-3}$ or more stops the build. On the four checkpoints, the submission's
  own forward pass reproduces the training states to within $2.0\times10^{-6}$. 
  
\item \textbf{Main Track} (Codabench 1010444): the submission archive is built from the
  Small Models Track archive by removing the track marker and adding the gpt-oss-120b branch of
  Algorithm~\ref{alg:probe}. All other files are identical.
  
\end{itemize}

\subsection{Reproduction}

The code is provided as supplementary material and is publicly accessible. Fitting the probe requires the stored prompt states and reasoning-length records, which the code regenerates on a GPU. The code fits the probe and its thresholds, and builds and checks both submission archives. It also reruns the MATH-split analysis, the probe variants, the steering experiment and the white-box ablation, and regenerates the result tables.

\section{Experimental details}
\label{app:data}

\paragraph{Data.} The official labeled split is \texttt{aimo-interp/train-main-v2} at revision
\texttt{23791d9b}~\citep{trainmainv2}. It has 82 rows over 36 problems, 71 of them labeled. A later revision
(\texttt{25e45665}) contained 7 labeled validation rows. It was not used in any analysis, and only its per-model row counts were inspected. The per-family results and the official perturbed texts come from
\texttt{aimo-interp-challenge-sample-full} at revision \texttt{c0ffb7e6}~\citep{samplefull}. The official texts
were used only as style examples for rewrite generation, and only from reference problems outside our labeled
set. Other data: Hendrycks MATH tables~\citep{hendrycks2021math}, a pool of 505 unlabeled olympiad problems
(OlympiadBench, AMC, and AIME 2024 and 2026), MathArena problem sets for the Main Track's recent problems,
ReliableMath (revision \texttt{c059d6e6})~\citep{xue2025reliablemath}, and MATH-Perturb,
Putnam-AXIOM and GSM-Plus for external checks~\citep{huang2025mathperturb,gulati2025putnam}. Because several
official problems repeat a template with one constant changed (36 problems over 29 templates), every evaluation
on the official split groups by template.

\paragraph{Data rules.} The organizers allowed additional labeled data, including data generated from the
official split.

\paragraph{Compute.} Small Models Track experiments ran on pairs of T4 GPUs on Kaggle. For the Main Track, we ran 37{,}632 base completions (seven configurations, eight samples, 672 problems) and 99{,}476
rewrite completions.

\section{Additional results}
\label{app:results}

\subsection{Hidden validation history}

\begin{figure}[H]
\centering
\includegraphics[width=\linewidth]{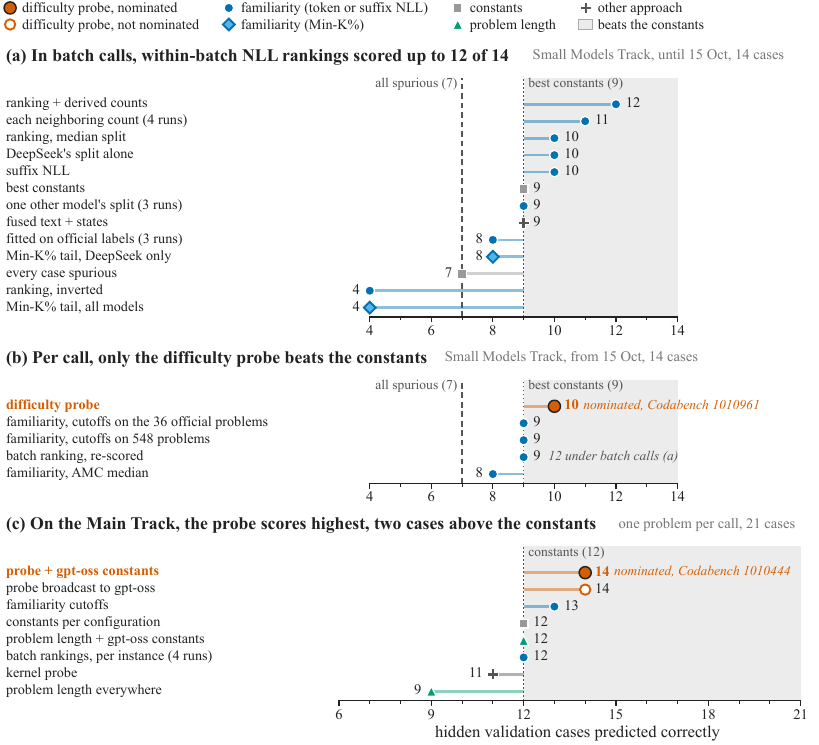}
\caption{Every scored submission used in this report. (a) Small Models Track under batch evaluation,
until 15 October. (b) The same track with one problem per call. (c) The Main Track with one problem per call. Each dot
is the number of hidden validation cases predicted correctly. The dotted line marks the best constants (9 of 14) and the Main
Track constants (12 of 21), the dashed line the all-spurious baseline (7 of 14), and shading the scores above the constants. A score below
chance (4 of 14) suggested that a familiarity ranking was informative but inverted, and its reverse scored 10 of 14.
Difficulty-probe runs are orange, and the nominated submissions are filled and carry their Codabench
IDs.}
\label{fig:ladder}
\end{figure}

Constant submissions measure each model's class balance on the hidden validation set. If predicting spurious
everywhere scores $a$, predicting robust for model $m$'s cases and spurious for all others scores $a + d_m$. On the Small Models Track, the
all-spurious submission scored 7 of 14 and the four single-model submissions scored 8, 8, 6 and 6 (Qwen,
DeepSeek, Skywork, Olmo) in each of two rounds, twelve days apart. Each submission scored under batch evaluation is a
deterministic function of one per-model ranking, so each score adds an equation in the same unknowns. Enumerating
all batch sizes and label patterns consistent with all 17 recorded scores leaves exactly one composition (5, 3, 3 and 3 cases for Qwen, DeepSeek, Skywork and Olmo, with 3, 2, 1 and 1 robust) and two label patterns that differ only within Qwen's batch. With
calibrated counts and a random ranking, the expected score is 8.05 when averaged over the 56 compositions
consistent with the balances. The probability of 12 or more is 0.055 (0.048 on the recovered composition).

\subsection{Main Track by configuration}

\begin{table}[H]
\centering\small
\caption{The nominated Main Track submission by configuration on the hidden validation set. The probe changes
two predictions relative to the constants, both correctly.}
\label{tab:main-config}
\setlength{\tabcolsep}{4.5pt}
\begin{tabular}{@{}lcccccc@{}}
\toprule
Configuration & Cases & Robust & Constant & Correct & Threshold & Correct\\
& & & & (constant) & & (submission)\\
\midrule
Qwen3.5-4B & 2 & 1 & robust & 1 & 10.386 & 1\\
Skywork-OR1-Math-7B & 3 & 1 & spurious & 2 & 9.969 & 3\\
Olmo-3-7B-Think & 2 & 1 & spurious & 1 & 9.944 & 2\\
DeepSeek-R1-0528-Qwen3-8B & 3 & 1 & robust & 1 & 10.392 & 1\\
gpt-oss-120b, low & 5 & 2 & spurious & 3 & -- & 3\\
gpt-oss-120b, medium & 3 & 2 & robust & 2 & -- & 2\\
gpt-oss-120b, high & 3 & 2 & robust & 2 & -- & 2\\
\midrule
Total & 21 & 10 & & 12 & & 14\\
\bottomrule
\end{tabular}
\end{table}

\begin{table}[H]
\centering\small
\caption{Main Track runs on the hidden validation set (Codabench logs). Every run had no invalid predictions and
no timeouts.}
\label{tab:main-runs}
\begin{tabular}{@{}llcc@{}}
\toprule
Run & Submission & Correct of 21 & Seconds\\
\midrule
Difficulty probe, gpt-oss constants (\textbf{nominated}) & 1010444 & 14 & 59.9\\
Same archive, second run & 1013643 & 14 & 59.0\\
Difficulty probe, Qwen's probe applied to gpt-oss & 1013026 & 14 & 122.2\\
Constants only & 991062 & 12 & 62.0\\
Problem length, gpt-oss constants & 1013902 & 12 & 1.9\\
Problem length for every configuration & 1013913 & 9 & 2.4\\
Familiarity thresholds, gpt-oss constants & 1010405 & 13 & 47.5\\
\bottomrule
\end{tabular}
\end{table}

\subsection{White-box ablation on all official rows}

\begin{table}[H]
\centering\footnotesize
\caption{The ablation on all 71 official rows (95\% intervals from 2{,}000 template resamples), where $n$
is the number of rows. ``Broadcast'' applies Qwen3.5-4B's probe to gpt-oss-120b's problems with one threshold per
effort. In ``all rows, nominated'', gpt-oss-120b uses its constants in every arm. Const., Probe and Length count correct rows, and the differences are in accuracy.}
\label{tab:ablation-all}
\setlength{\tabcolsep}{2.5pt}
\begin{tabular}{@{}lcccccc@{}}
\toprule
Rows & $n$ & Const. & Probe & Length & Probe $-$ const. & Probe $-$ length\\
\midrule
Small models & 34 & 18 & 28 & 24 & $+0.294$ [$+0.062$, $+0.519$] & $+0.118$ [$+0.024$, $+0.259$]\\
gpt-oss-120b, broadcast & 37 & 25 & 26 & 27 & $+0.027$ [$-0.205$, $+0.231$] & $-0.027$ [$-0.194$, $+0.118$]\\
All rows, nominated & 71 & 43 & 53 & 49 & $+0.141$ [$+0.033$, $+0.258$] & $+0.056$ [$+0.013$, $+0.121$]\\
All rows, broadcast & 71 & 43 & 54 & 51 & $+0.155$ [$-0.014$, $+0.312$] & $+0.042$ [$-0.048$, $+0.149$]\\
\bottomrule
\end{tabular}
\end{table}

Applied to gpt-oss-120b's problems, Qwen3.5-4B's probe ranks the 37 official gpt-oss rows at an
area under the receiver operating characteristic curve (AUC) of 0.853
[0.663, 0.980], but problem length ranks them almost as well (0.762 [0.553, 0.918]). The probe uses one
threshold per effort, set from the hidden set's robust share in each configuration and fixed before any score was
computed. Its decisions are correct on 6 of 7 low-effort rows, 11 of 15 medium and 9 of 15 high, against 2, 10 and
13 for the constants. On the 19 rows whose problem has no labeled small-model row, its AUC is 0.648 and it is correct
on 8 rows against the constants' 11. On the hidden set, its two changed predictions cancel.

\subsection{The relation by population}

\begin{table}[H]
\centering\small
\caption{Values plotted in Figure~\ref{fig:direction}a, as AUC with 95\% intervals. $\dagger$: subset
selected post hoc. The official-split familiarity value has no recorded interval.}
\label{tab:fig2a}
\setlength{\tabcolsep}{4pt}
\begin{tabular}{@{}lccc@{}}
\toprule
Stated direction & Official split (34) & MATH split (137) & \begin{tabular}[b]{@{}c@{}}MATH, \pfam{domain}\\never applied (39)$^\dagger$\end{tabular}\\
\midrule
Easier $\Rightarrow$ robust (probe) & 0.883 [0.706, 1.000] & 0.346 [0.246, 0.447] & 0.672 [0.492, 0.842]\\
Familiar $\Rightarrow$ robust (token NLL) & 0.358 & 0.683 [0.582, 0.780] & 0.506 [0.311, 0.700]\\
\bottomrule
\end{tabular}
\end{table}

\subsection{What the probe reads}

\begin{table}[H]
\centering\small
\caption{Probe variants (exploratory). Each row refits the probe unchanged except as stated.
TF-IDF: term frequency--inverse document frequency.}
\label{tab:probe-variants}
\begin{tabular}{@{}lcc@{}}
\toprule
Variant & CV Spearman (unlabeled) & AUC (official split)\\
\midrule
All layers, averaged (submitted) & 0.768--0.788 & 0.883\\
Single layer, first tenth of depth & 0.694 & 0.812\\
Single layer, half to three quarters depth & 0.782 & 0.904\\
Top 1 / 5 / 20 / 50 principal components & 0.450 / 0.631 / 0.753 / 0.758 & 0.554 / 0.758 / 0.858 / 0.879\\
Twelve surface features & 0.523 & 0.700\\
Word TF-IDF, ridge on the same target & 0.603 & 0.871\\
\bottomrule
\end{tabular}
\end{table}

The four models' predictions agree at pairwise Spearman correlations of 0.926 or higher (mean 0.956) on the 51
problems outside the fit. Surface features explain the probe's score with an $R^2$ of 0.48--0.50, and the
unexplained part alone still ranks the official split at an AUC of 0.642. The TF-IDF model ranks the official
split almost as well as the probe, so the 34 official rows cannot separate the two. It fits held-out olympiad
problems much worse (0.603 against 0.78) and agrees with the probe at only 0.546, so the 497 unlabeled problems favor
the hidden states. Long predicted reasoning goes with answer formats and enumeration
(``irreducible fraction'', ``1000'', ``least''), and short predicted reasoning with direct computation
(``compute'', ``what is''). Refitting the probe on random subsets of its training problems changes its
official-split AUC by about 0.08 (0.815 $\pm$ 0.082 with 125 problems), so the official split cannot distinguish
refits of one method. A refit on a target with a 64k-token cap reduced the share of problems with both
samples capped from 28\% to 9\%. It did not improve the official-split AUC (0.808 against 0.883) and was not
submitted.

\begin{table}[H]
\centering\small
\caption{Steering along the probe direction (pre-registered). Qwen3.5-4B, layer 15 (depth 0.47).
The push of $2\,\mathrm{sd}$ is added to every prompt token during prefill, where $\mathrm{sd} = 0.0719$ is the standard deviation of the prompt states
along the direction. The test used 60 unlabeled problems,
two samples per condition and a 16k-token cap.}
\label{tab:steering}
\begin{tabular}{@{}lcc@{}}
\toprule
Intervention, relative to none & Mean log reasoning length & Probe reading, layers 16--32\\
\midrule
$+2\,\mathrm{sd}$ along the probe direction & $-0.003$ [$-0.114$, $+0.111$] & $+0.187$\\
$-2\,\mathrm{sd}$ along the probe direction & $-0.004$ [$-0.122$, $+0.113$] & $-0.186$\\
$+2\,\mathrm{sd}$ along a random direction & $+0.062$ [$-0.047$, $+0.167$] & $+0.012$\\
\bottomrule
\end{tabular}
\end{table}

If reasoning length followed the probe's reading one for one, the difference between the
$+2\,\mathrm{sd}$ and $-2\,\mathrm{sd}$ conditions would be 0.26--0.37 log-tokens. The observed difference is
$+0.001$ [$-0.099$, $+0.111$].
Accuracy (0.93--0.96) and the share of capped samples (7.5--9.2\%) are flat across conditions, so the
intervention did not damage the model.

\subsection{Contradiction detection by familiarity}
\label{app:contradiction}

\begin{table}[H]
\centering\small
\caption{Contradiction detection by familiarity (pre-registered). Qwen3.5-4B on 177 olympiad problems it solved
twice. The study used two verified contradiction variants per problem, four samples each, a prompt that permits
declining and a 32k-token cap.}
\label{tab:contradiction}
\begin{tabular}{@{}lccc@{}}
\toprule
NLL quintile (1 = most familiar) & Problems & Detection rate & Capped samples\\
\midrule
1 & 36 & 0.229 & 0.399\\
2 & 35 & 0.250 & 0.361\\
3 & 35 & 0.336 & 0.393\\
4 & 35 & 0.464 & 0.296\\
5 & 36 & 0.389 & 0.236\\
\bottomrule
\end{tabular}
\end{table}

Table~\ref{tab:contradiction} gives the detection rate by familiarity quintile. The Spearman correlation between mean
token NLL and detection rate is 0.256 [0.112, 0.386]. Within length
terciles it is 0.142 [$-0.015$, 0.282], and over untruncated samples only it is 0.213 [0.069, 0.357]. The study
uses one model and contradictions written by another model, so it supports the proposed mechanism
(a model that recognizes a problem checks an added premise less carefully) without establishing it.

\clearpage
\section{Rejected hypotheses}
\label{app:negatives}

\begin{table}[H]
\centering\footnotesize
\caption{Hypotheses we tested and rejected, the strongest result each reached, and the test that rejected it.}
\label{tab:negatives}
\setlength{\tabcolsep}{3pt}
\begin{tabular}{@{}r>{\raggedright\arraybackslash}p{4.25cm}>{\raggedright\arraybackslash}p{3.5cm}>{\raggedright\arraybackslash}p{4.65cm}@{}}
\toprule
& Hypothesis & Strongest result & Rejected by\\
\midrule
1 & Linear probes on hidden states predict robustness & 58.4\% as published, AUC 0.75--0.85 within folds & Fits average 0.501 per layer, and nested selection gives 0.552\\
2 & A model trained on MATH transfers to olympiad problems & 0.692 balanced accuracy in distribution & AUC 0.429 on olympiad rows\\
3 & Training on the official labels instead rescues it & -- & Not at 33 labeled problems\\
4 & Self-labeling on MATH reproduces the official labels & -- & Cohen's $\kappa = -0.135$\\
5 & Self-labeling on olympiad problems does & $\kappa = 0.286$ & Interval spans zero, and about one label per GPU-hour\\
6 & Answer agreement under rewriting & -- & AUC 0.458 at 28k generated tokens per problem\\
7 & Reasoning-trace uncertainty features & -- & Near chance, and trace states require full generation\\
8 & Architecture-invariant activation geometry & Family-wise selected signal & Reverses sign on independent MATH data\\
9 & Counterfactual prompt-state sensitivity & AUC 0.735 & 0.545 [0.428, 0.658] on an independent population\\
10 & External robustness labels transfer & Real labels, 233 configurations & Putnam-AXIOM AUC 0.609 [0.442, 0.769]\\
11 & Budgeted reasoning diagnostics & Ranking gates and an independent confirmation passed & 7{,}526\,s for two models against a fixed budget of 3{,}000\,s\\
12 & Min-K\% tail statistic, best of twelve offline & AUC 0.867 [0.728, 0.973], maxT $p = 0.001$ & 4 of 14 on the hidden set\\
13 & Publication status explains the familiarity reversal & -- & Both halves reverse (AUC 0.417 and 0.299)\\
14 & The reversal survives a continuous target & Spearman $+0.247$ & Every interval spans zero\\
15 & Budgeted diagnostics in the evaluation pipeline & Problem-level AUC 0.659 & Balanced accuracy 0.521 against a bar of 0.55\\
16 & One pass predicts answering through an impossible premise & -- & No positive class (2 of 130 and 0 of 274 flagged)\\
17 & A held signal beats the constants per instance & Putnam-AXIOM 0.598, GSM-Plus 0.564 & Labels every official row spurious, and 60 calls need at least 496 minutes\\
18 & Familiarity as a per-instance threshold & MATH split 0.624 [0.545, 0.697] & 8, 9 and 9 of 14 against the constants' 9\\
19 & Easier $\Rightarrow$ robust beyond the official split & AUC 0.883 official, and 10 of 14 hidden & AUC 0.346 [0.246, 0.447] on the MATH split\\
20 & A target censored at 64k ranks the official split better & Censoring 28\% $\to$ 9\% & AUC 0.808 against 0.883, and 26 against 28 of 34\\
\bottomrule
\end{tabular}
\end{table}

\section{Self-labeling for the Main Track}
\label{app:labels}

To obtain labels at scale, we applied the organizers' procedure ourselves. One generator wrote candidate
rewrites in the four model-written official families. Typos were applied mechanically. A panel of three
other model families re-solved each candidate. A candidate was kept only if every panel model's majority answer
matched the problem's answer. Labels then followed Equation~\ref{eq:label}.

\begin{table}[H]
\centering\small
\caption{Rewrite survival against a pre-registered bar of 0.90. Survival is the number confirmed divided by the number verified.}
\label{tab:survival}
\begin{tabular}{@{}lccc@{}}
\toprule
Family & Verified & Confirmed & Survival\\
\midrule
\pfam{rephrase} & 465 & 410 & 0.882\\
\pfam{distract} & 480 & 394 & 0.821\\
\pfam{rename} & 422 & 326 & 0.773\\
\pfam{domain} & 351 & 260 & 0.741\\
\bottomrule
\end{tabular}
\end{table}

We labeled 507 (problem, configuration) pairs over 96 problems, 370 of them robust. Only 37 problems received
the same label in every configuration, and a problem's majority label matches 0.826 of our labels against
0.986 of the official ones. Against the official labels, leave-one-template-out agreement was
Cohen's $\kappa = 0.200$ [0.000, 1.000] on 24 rows, against a pre-registered bar of 0.4. The Spearman correlation between our largest accuracy drops $\Delta$ (Equation~\ref{eq:label}) and the official ones was 0.224
[$-0.128$, 0.619] on 25 rows. All three pre-registered conditions failed, and none of eight rankings
evaluated on these labels had an interval above 0.5. Much of the survival shortfall appears to come from the
panel rather than the rewrites. On the calibration problems, 44 candidates failed only because one panel
model's most frequent answer was missing or tied. These failures do not distinguish noisy labels from absent signal.

\end{document}